\documentclass[a4paper, 10pt, conference]{IEEEtran}
\IEEEoverridecommandlockouts
\usepackage{cite}
\usepackage{enumitem}
\usepackage{amsmath,amssymb,amsfonts}
\usepackage{algorithmic}
\usepackage{graphicx}
\usepackage{textcomp}

\usepackage{xcolor}
\usepackage[ruled,vlined,linesnumbered]{algorithm2e}
\SetKwRepeat{Do}{do}{while}%
\def\BibTeX{{\rm B\kern-.05em{\sc i\kern-.025em b}\kern-.08em
    T\kern-.1667em\lower.7ex\hbox{E}\kern-.125emX}}
\begin{document}

\title{On Decentralizing Federated Reinforcement Learning in Multi-Robot Scenarios\\
}

\author{\IEEEauthorblockN{Jayprakash S. Nair}
\IEEEauthorblockA{\textit{Dept. of Computer Science and Engg.,} \\
\textit{Federal Institute of Science and Technology,}\\
Angamaly, India \\
jsnair.hi@gmail.com}
\and
\IEEEauthorblockN{Divya D. Kulkarni}
\IEEEauthorblockA{\textit{Dept. of Computer Science and Engg.,} \\
\textit{Indian Institute of Technology Guwahati,}\\
Guwahati, India \\
divyadk@iitg.ac.in}
\and
\IEEEauthorblockN{Ajitem Joshi}
\IEEEauthorblockA{\textit{Dept. of Computer Science and Engg.,} \\
\textit{Indian Institute of Technology Guwahati,}\\
Guwahati, India \\
j.ajitem@iitg.ac.in}
\and
\IEEEauthorblockN{Sruthy Suresh}
\IEEEauthorblockA{\textit{Dept. of Computer Science and Engg.,} \\
\textit{Federal Institute of Science and Technology,}\\
Angamaly, India \\
sruthyajith@gmail.com}
}

\maketitle

\begin{abstract}
Federated Learning (FL) allows for collaboratively aggregating learned information across several computing devices and sharing the same amongst them, thereby tackling issues of privacy and the need of huge bandwidth. FL techniques generally use a central server or cloud for aggregating the models received from the devices. Such centralized FL techniques suffer from inherent problems such as failure of the central node and bottlenecks in channel bandwidth. When FL is used in conjunction with connected robots serving as devices, a failure of the central controlling entity can lead to a chaotic situation. This paper describes a mobile agent based paradigm to decentralize FL in multi-robot scenarios. Using Webots, a popular free open-source robot simulator, and \textit{Tartarus}, a mobile agent platform, we present a methodology to decentralize federated learning in a set of connected robots. With Webots running on different connected computing systems, we show how \textit{mobile} agents can perform the task of Decentralized Federated Reinforcement Learning (dFRL). Results obtained from experiments carried out using Q-learning and SARSA by aggregating their corresponding Q-tables, show the viability of using decentralized FL in the domain of robotics. Since the proposed work can be used  in conjunction with other learning algorithms and also real robots, it can act as a vital tool for the study of decentralized FL using heterogeneous learning algorithms concurrently in multi-robot scenarios. 

\end{abstract}

\begin{IEEEkeywords}
Decentralized Federated Learning, Reinforcement Learning, Networked Robots, Mobile Agents
\end{IEEEkeywords}









\section{Introduction}
With an ever increasing trend in the use of handheld devices and a consequent enormous explosion in data generated, researchers have been trying hard to figure out varied techniques to learn from such data by aggregating the same in a centralized entity. The learning algorithm is then run on this central server and the knowledge gained is sent back to all connected participating devices. This process is not always viable considering the fact that a large amount of data needs to be uploaded to the server and then processed periodically \cite{mcmahan2021advances}. One work-around is to use a technique  termed  Federated  Learning  (FL) \cite{mcmahan2017communication} where,  in  lieu  of  data,  the  models generated \textit{in-situ} or on-device are shared by all participating devices with the central server. The server, in turn, aggregates them using some pre-defined techniques \cite{abdulrahman2020survey, konevcny2016federated, wang2017co} and sends the modified model back to the devices. Thus, every device has a local dataset using which, a model, most often an Artificial Neural Network (ANN), is trained and evolved. Each device periodically sends the trained weights of its respective ANN to the central server in the form of an update. At the server these weights are, for instance, averaged, and then sent back to all the devices. Most often these new set of (averaged) weights represent a part of the learning performed at each of the participating devices and hence contribute to the enhancement of learning within the network. Over several rounds of this process, the models at each of the devices saturate to fairly homogeneous ones. This centralized version of FL model suffers from inherent drawbacks such as a central point of failure, scalability, privacy issues, coupled with the requirement of large clients-to-server bandwidth \cite{mcmahan2021advances}. In order to overcome these hurdles, decentralized versions of FL have been proposed \cite{lalitha2018fully, hu2019decentralized,roy2019braintorrent, pokhrel2020decentralized, agrawal2020decentralizing}.
FL has also made its niches in multi-robot scenarios \cite{xianjia2021federated, majcherczyk2021flow}. In such cases, each robot shares its learned model with others, thereby aiding in faster learning convergence. The robots could be either in the same environment or different ones. Robots, most often need to connect to a centralized server, a cloud or a controller, which in turn performs the task of aggregating the models received. Since robots could be mobile, their dynamically changing positions may tend to make or break connections with the central entity. For a group of mobile robots in the same environment or different environments, a decentralized approach or a hybrid of the centralized and decentralized approaches, could prove to be more beneficial.  Research on FL in the area of robots, most often target  specific or customised robotic scenarios, making it difficult for others to reuse the work. 

This paper describes a mobile agent based decentralized technique to allow robots simulated within several instantiations of Webots \cite{michel2004cyberbotics} running on different connected machines to use a learning algorithm in a federated manner.  
A mobile agent, knitting through these Webots instantiations, serves to collate, aggregate and share the learned models of the learning algorithms running concurrently on each of the robots in the separate, yet connected Webot environments, thereby facilitating FL amongst the robots in a decentralized manner. In the experiments performed, we have used two variants of Reinforcement Learning (RL) \cite{sutton2018reinforcement}, viz. Q-Learning and State-Action-Reward-State-Action (SARSA) \cite{SARSA}, to make the robots, inhabiting different environments, learn obstacle avoidance behaviour using Decentralized Federated Reinforcement Learning (dFRL). However, it may be noted that in lieu of RL, other learning algorithms could be used in a decentralized and federated manner, using the proposed mobile agent based method.  By merely replacing the learning controller of one robot by another, this generic method can cater to a set of robots using heterogeneous learning algorithms.
Subsequent sections describe RL, the concept of mobile agents and platforms, Webots and mechanisms to realize Decentralized Federated RL (dFRL) followed by experiments conducted and results obtained, using Webots running concurrently on different networked machines. The paper also enumerates the different ways in which the proposed work could be used in the section on Discussions and Implications before concluding.

\section{Reinforcement Learning}
Learning algorithms may be broadly classified into Supervised, Unsupervised and Reinforcement based. While the first which nominally uses a pre-defined set of rules or labelled datasets that provide the answers to learn and generalize inputs and outputs, the second tries to learn to conclude from unlabelled data by extracting features and patterns on its own. For the past few years there has been a resurgence of the use of Reinforcement Learning (RL), especially in the domain of robotics \cite{kober2013reinforcement, mataric1997reinforcement}. RL can serve as a powerful tool in robotics to facilitate autonomous discovery through trial-and-error interactions with the environment. Several researchers have tried their hand at using Q-Learning \cite{watkins1992q}, which is one of the more prominent types of RL \cite{sutton2018reinforcement} approaches. Most of the work on Q-Learning used in robotics, describe attempts to learn lower-level behaviours, such as obstacle avoidance, wall following, etc., in a given environment. Though behaviours such as obstacle avoidance may seem rudimentary and thus are programmed \textit{a priori}, they may not work the same way as envisioned in the real world. In dynamic environments and also changing robot parameters, such programs may eventually fail to produce proper results in real world conditions. It is thus, vital to embed mechanisms to learn even low level behaviours within a robot so that they can prove to be useful even in new or unknown terrains.
In RL, an agent in the initial state $S$, performs an action $A$ by virtue of which it transits to another state $S’$ in its environment. In doing so the agent is provided a reward based on the goodness of the action performed. The algorithm stores the information about the initial state, the action performed and the reward, for future use. In the new state, the agent selects and performs another action and correspondingly gets another reward. This process continues till it finally reaches its designated goal. The agent performs several such episodes to eventually learn the best set of actions to reach the goal. 



Some terms pertaining to RL are briefly defined below:
\begin{itemize}

    \item Action (A): It constitutes the possible moves that the agent can make.
   \item State (S): It is the current situation returned by the environment.
    \item Reward (R): It comprises an immediate return received from the environment to evaluate the last action performed by the agent.
   \item Policy ($\pi$): It is strategy that the agent employs to determine the next action based on the current state.
 \item Q-value: This value provides an estimation of how good it is to take the action in a given state. It refers to the long-term return of the current state s, taking action a under the policy, $\pi$.

\end{itemize}

Q-Learning is one type of RL which has been found to be well suited to cater to real environments and hence is widely used in the domain of robotics. 

\subsection{Q-Learning}

The behaviour of the agent or robot in this case is defined by its policy. This policy could be implemented in different ways based on the method. It could be in the form of a table, a function or even an Artificial Neural Network. Tabular Q-learning, which is the most basic version, uses a table, wherein there is a row and a column designated for each action. This table informs the agent of the expected result of performing an action in the given state. When the state of the environment changes, the agent checks the row corresponding to the given state and chooses the action that in the past yielded the highest reward. The values for each action and state are referred to as the Q-values.


Initially, every Q-value in the table is initialised to 0. The values are then updated after each step using an update rule which uses the reward received after an action is performed. A reward is a measure of how good the new state is. The Q-Learning update rule is given below:

\begin{multline}
     Q_{n}(s,a) = Q(s,a) + \alpha*[ R_{n}(s,a) + \\ (\gamma* Q_{max_a}(s’,a')) - Q(s,a)]
\end{multline}
  
where $Q_{n}(s,a)$ and $Q(s,a)$ stand for the new and old Q-values of the earlier state, $Q_{max_a}(s,a)$ is the maximum of the Q-values considering the new state the agent is now in (after performing the action $a$), $r$ is the reward the agent received, $\alpha$ is the learning rate and $\gamma$ is the discount factor. 
\newline
The agent repeats the episodes several times, always updating its policy after each step using its new experience (state, action, new state, reward). Eventually, the agent learns the policy that yields a good reward over the episode, just as a person would learn the manner in which he/she has to play a game to obtain a good score.

\subsection{State-Action-Reward-State-Action (SARSA)}
SARSA \cite{sutton2018reinforcement} is a variant of Q-learning, which in turn is a form of RL. Unlike Q-learning, SARSA is of the on-policy type and  uses the action performed by the current policy to learn the Q-value. The updates in SARSA occur based on the equation below:
\begin{multline}
    Q_{new}(s,a) = Q_{old}(s,a) + \\ \alpha*[r + \gamma * Q(s’,a’) - Q_{old}(s,a)]
\end{multline}


where $s’$ and $a’$ form the next state and next action. As can be observed, the update is dependent on the current state, action, reward, and the next state and action, i.e. the tuple (s,a,s’,a’). This is the main reason why it is called State-Action-Reward-State-Action or SARSA. 

A common yet vital behaviour, that needs to be learnt by an autonomous robot inhabiting different heterogeneous terrains, is obstacle avoidance. Obstacle avoidance scenarios could be of two types – (i) when we have the complete knowledge of the environment, and (ii) when this knowledge is either partially known or is totally unknown. In the former case where the environment appears to be deterministic, classical methods \cite{masmoudi2016fuzzy,pan2019multi} could be used to exhibit this behaviour. Unfortunately, since the real world exhibits a host of non-deterministic events, the efficiency of such methods rapidly decreases with the complexity and the unstructured nature of the environments. Implementing classical methods also entail a considerable amount of modelling. RL, can come to the aid in such situations, since it does not require a complete knowledge of the environment. In addition, it also empowers the robot with an online learning capability. In RL, the designer of the controller provides feedback in the form of a scalar objective function that provides a measure of the one-step performance of the robot. This paper, thus describes how Q-learning and SARSA, can be used for learning obstacle avoidance in a decentralized and federated manner by a set of robots.

\section{\textit{Tartarus} - A Mobile Agent Platform}
An agent is a software that can autonomously make a decision that can be put into action. It senses its environment and then performs the requisite operation. Its  actuators could be used to initiate the action in its environment. Robots may be looked upon as their hardware counterparts. Software Agents could be either $static$ or $mobile$. While the former is a software that remains resident on a computing system or node, the latter has the capability to move from one computing node in a network to another connected one and execute its associated operations or behaviours. Naturally, every computing node would need a platform to host such agents, resources for the agents to execute and ensure their eventual dispatch to another node. Several such agent platforms exist \cite{wong1997concordia, bellifemine1999jade, chenagent, Aglets, Typhon}. \textit{Tartarus} \cite{tartarus} is one such open source platform, which is written in SWI-Prolog. It provides for creating, programming, and empowering agents with a range of features including mobility, cloning and payload carrying ability, etc. These agents can be created at a node and can be set free to migrate in a network of such nodes, each of which hosts a \textit{Tartarus} platform. The platform can be run on  operating systems such as  Windows, Linux and Raspbian. This allows it to be run on embedded controllers such as the Raspberry Pi and access all its hardware features. \textit{Tartarus} also comes with add-ons for controlling robots such as Lego Mindstorm NXT and Firebird. The work described in this paper uses \textit{Tartarus} for implementing all mobile agent functionalities.

\section{Webots}
Webots \cite{michel2004cyberbotics} is an open source multi-platform  application used to simulate different types of simulated robots. It facilitates creation and modelling of the environment or arena inhabited by the robots, and also allows programming of their respective controllers. It provides access to sensors, actuators and objects within the simulated environment. Its Graphical User Interface (GUI) comes with both physics and rendering engines that allow for immaculate simulations and programming of robot controllers. Webots can be run on Windows, Linux and Mac OS and the controllers within can be programmed using C, C++, Python, Java, MATLAB or \textit{Robot Operating System (ROS)}. We have used Webots to simulate a multi-robot scenario, and demonstrate the use of mobile agents to aid decentralized federated reinforcement learning amongst the robots. 

\section{Decentralized Federated Reinforcement Learning Methodology}
In order to ensure that the simulated robots within Webots running across connected machines share their respective learned models in a decentralized and federated manner, we have made use of mobile agents. Mobile agents exhibit all characteristics of conventional static agents but in addition can actually move from one computing node in a network to another and execute the actions they are empowered to, at the destination node. Both static and mobile agents require the existence of an agent platform at every node in the network.  We have used the \textit{Tartarus} agent platform, to program and facilitate agent mobility. Being autonomous and mobile, a mobile agent in \textit{Tartarus} can carry the model learnt by one robot in the network, as payload, to another in the same network thereby facilitating the sharing of the models between the networked robots. Since they function without the need of a central controlling entity, they help in achieving decentralized sharing of the information. In addition, such agents can also be made to process the models they carry, based on the models generated at the destination robot. These agents can thus, act as model aggregators and consequently hasten the process of learning. For instance, having acquired a (partially) learned model $M_1$, from the robot $R_1$, a mobile agent $A_1$ can migrate to another connected robot $R_2$ (running on a different system), procure its current model, $M_2$, average the two models, and provide the aggregated one, $M$, to $R_2$. The agent can then migrate to another connected robot $R_3$ and perform a similar job, thereby effecting federated learning in a decentralized manner. In this work, the robots simulated within Webots running on different connected computers use either Q-learning or SARSA in their respective environments to learn to avoid obstacles within. Alongside Webots, a \textit{Tartarus} platform running on each of these connected systems, facilitates agent mobility and execution. A mobile agent is programmed in such a way that it migrates to one such system, connects to the robot within, and copies the Q-table learned by that robot. It carries this table as its payload and migrates to another connected system running Webots wherein it aggregates this Q-table with the one locally available and migrates to another such system. Once all such systems have been visited the Q-table carried by it constitutes the aggregate of the Q-tables of all the robots. The agent then retraces its path and provides the aggregated version of the table to all robots \textit{en route}. Having replaced all the Q-tables at all systems with the aggregated version, the mobile agent repeats the process again after a brief respite. This respite period, allows for robots, to find the efficacy of the aggregated version in their respective environments and populate new values within.
As can be inferred this Decentralized Federated Reinforcement Learning (dFRL), performed by the mobile agent, over several such migrations across these connected robots, results in a better Q-table.
Algorithm \ref{algo} depicts the algorithm used to achieve dFRL. Initially, the learning process is commenced in all the robots, $R_1,R_2,..,R_n$, operating within their respective Webots instantiations. For the first $m$ iterations (respite period), the respective Q-tables are let to build-up and be populated, after which these tables are aggregated and provided to all the robots by the mobile agent. This process is made to continue till the learning saturates.

\begin{figure}[!t]
    \centering
    \includegraphics[height=4.6cm, width=6cm]{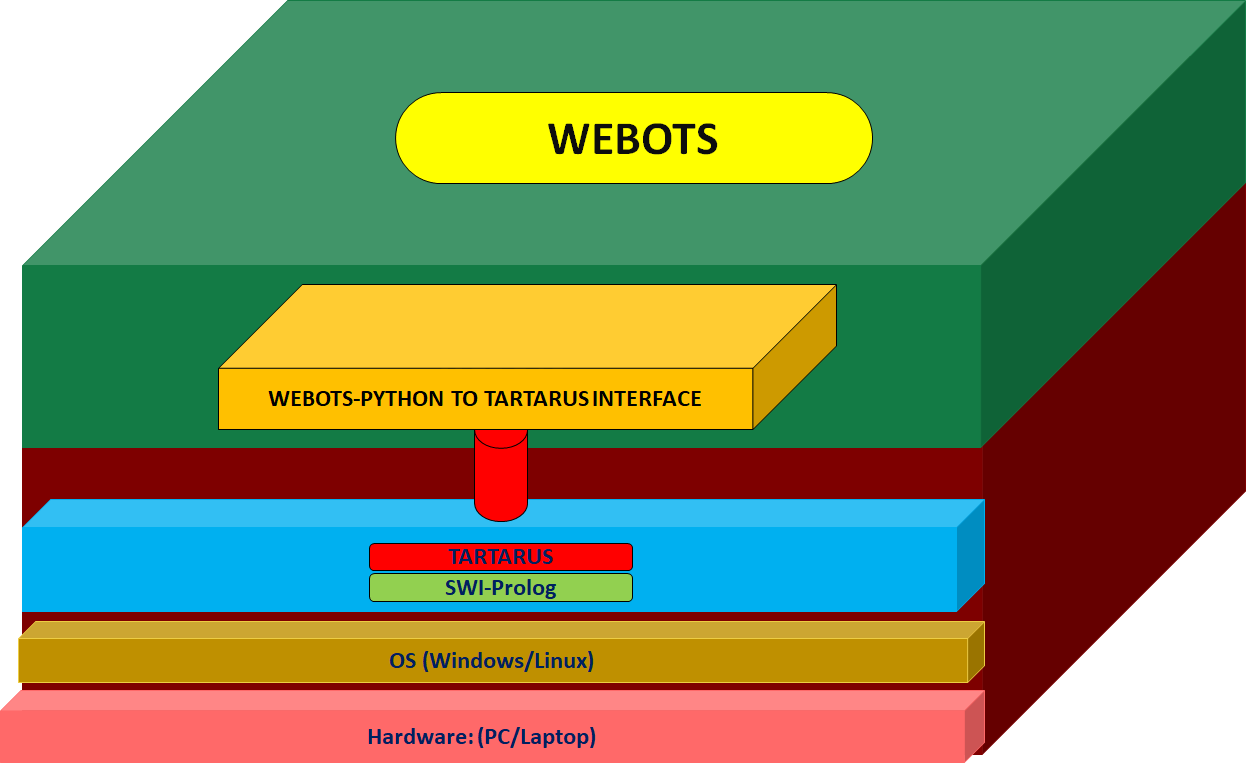}
    \caption{Architectural support for Agents within Webots}
    \label{architecture}
\end{figure}

\begin{figure}[!t]
    \centering
    \includegraphics[height=4.6cm, width=6cm]{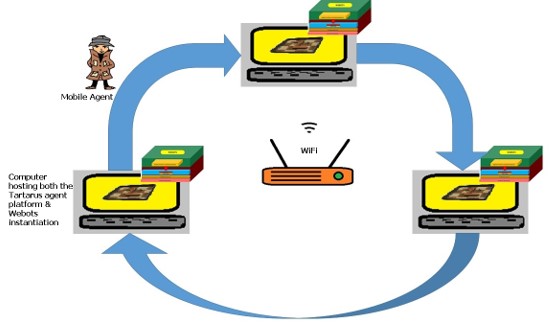}
    \caption{Decentralized Federated Learning environment using Webots}
    \label{env}
\end{figure}

\section{Mobile Agent based Decentralized Federated Learning Architecture}
Fig.\ref{architecture} shows the architecture used to supports mobile agents using \textit{Tartarus} in a single Webots environment. As can be seen, the bottom two layers comprise the hardware and the operating system. Since \textit{Tartarus} has been written solely using SWI-Prolog \cite{wielemaker2012swi}, these two layers together occupy the next upper layer. Since the robot controller within Webots was programmed in Python (running from within Webot), it was essential for us  to realize a \textit{Tartarus}  to \textit{Webots} interface so as to bridge the controller with the mobile agent hosted within \textit{Tartarus}. This interface, which forms the next layer, makes use of a function named $PySwip$, to bridge Python with SWI-Prolog. The bridge facilitates querying SWI-Prolog via Python. The \textit{Webots-Python} to \textit{Tartarus} bridge serves as a seamless interface that can perform all required transactions between the simulated robots and the \textit{Tartarus} based mobile agent. It converts all Python queries into corresponding \textit{Tartarus} commands to facilitate agent programming from within Webots. Using this interface, Webots programmers can now include agent programming paradigms into their respective Webots-Python code. Since both Webots and \textit{Tartarus} can be interfaced to real robots, the realization of this interface opens up a plethora of applications catering to both soft and hard agents in the areas of distributed and decentralized control and learning. This layer, empowers Webots programmers to use all features of \textit{Tartarus} and integrate the simulated and real robots with static and/or mobile agents.

\begin{algorithm}[t]
\KwData{n the number of robots, learning algorithm}
\KwResult{Aggregate Q-table}
init\_learning($R_1,R_2,R_3,...,R_n$) \\
\Do{end of learning}{
      At the end of $m$ iterations by $R_0$: \\
      aggreg\_Q=$R_0^Q$ \\
      \For{i$\gets$2 to n}{
        $R_i$ $\gets$ $send(aggreg\_Q)$ \\
        aggreg\_Q = $aggregate(aggreg\_Q, R_i^Q)$ \\
      }
      $R_{n}^Q$ = $aggreg\_Q $ \\
      \For{i$\gets$n-1 to 1}{
        $R_{i}$ $\gets$ $send(aggreg\_Q)$ \\
        $R_{i}^Q$ = $aggreg\_Q$
      }
    }
\SetAlgoLined

\caption{The proposed dFRL algorithm}
\label{algo}
\end{algorithm}

Fig.\ref{env} depicts a typical Decentralized Federated Learning environment with Webots and the agent architecture running on different computing systems forming a network. While Webots runs on each of the connected machines alongside  the agent platform, the agent connects to the corresponding robot via the interface, procures the locally available model, performs the aggregation and continues to migrate to the next machine with the new aggregated model based on the procedure described in Algorithm 1. Thus, robots and agents communicate with one another via the interface which in turn facilitates both accessing and modifying the respective models (Q-tables).

\begin{figure}[!t]
    \centering
    \includegraphics[height=4.6cm, width=6cm]{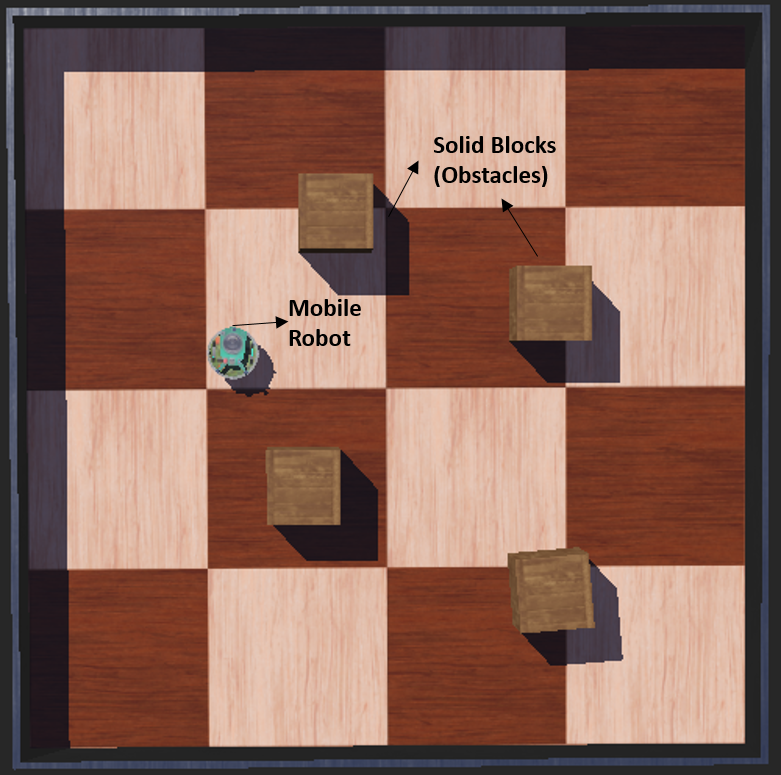}
    \caption{Arena of the Webots with the robot $R_4$ and the solid blocks}
    \label{q4arena}
\end{figure}

\begin{figure}[!t]
    \centering
    \includegraphics[height=4.6cm, width=6cm]{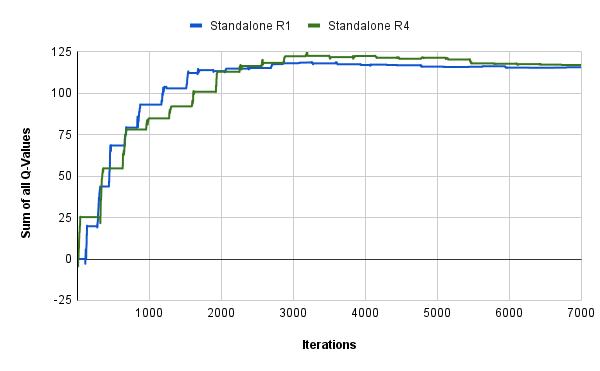}
    \caption{Sum of all Q-Values over iterations of the Q-Learning on standalone (Non-Federated) robots $R_1$ and $R_4$}
    \label{standaloneq}
\end{figure}

\begin{figure}[!t]
    \centering
    \includegraphics[height=4.6cm, width=6cm]{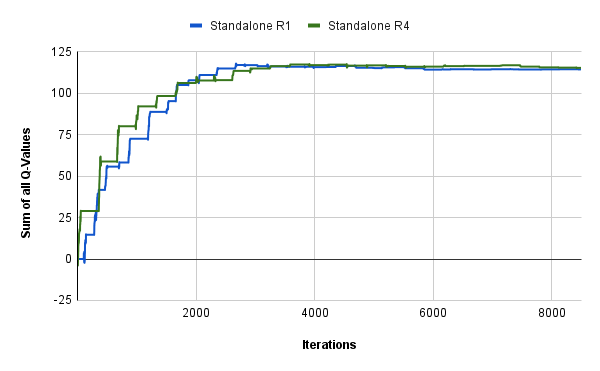}
    \caption{Sum of all Q-Values over iterations of the SARSA on standalone (Non-Federated) robots $R_1$ and $R_4$}
    \label{standalonesarsa}
\end{figure}

\section{Experiments and Results}

In the actual experimentation to test the mobile agent based dFRL, we used five identical personal computers connected via a LAN, each running a \textit{Tartarus} agent platform and one Webots instantiantion with a robot within its arena. All the environments (Webots arenas) were dissimilar. The first arena with robot $R_1$ had a single solid block acting as an obstacle, while those of $R_2$ and $R_3$ had two such blocks placed randomly. The arena of $R_4$ had four such blocks (shown in Fig. \ref{q4arena}). The arena of $R_5$ had no such blocks. In all the cases, the boundaries of the arenas also acted as obstacles.
The use of different arenas in each of the Webots instantiations essentially meant that all robots tackled different environments. This was done to ensure that the models learnt by individual robots from their respective environments (arenas), were different. The federated learning could thus, aid in aggregating these diverse models learnt from different arenas, and then share the same with all participating robots. The \textit{Tartarus} platform running alongside the Webots instantiation in a computer facilitated the hosting of the mobile agent which in turn aided in decentralizing the federated learning process. As mentioned earlier, for learning, we used two types of RL techniques viz. Q-learning and SARSA. These were coded to make the robots learn the obstacle avoidance behaviour.
Averaging and maximum of Q-values, were used as the aggregation methods, in both the Q-learning and SARSA based federated learning experiments. While in the former, the Q-values within the two tables were averaged, in the latter, the Q-values of two tables were compared and the maximum of the corresponding values were used to populate the aggregated Q-table.

\begin{figure}[!t]
    \centering
    \includegraphics[height=4.6cm, width=6cm]{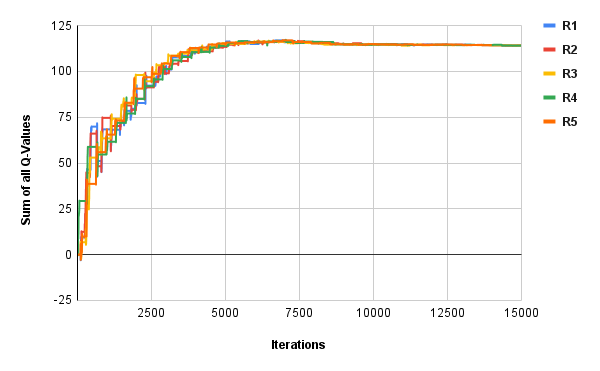}
    \caption{Sum of all Q-Values over iterations of the Federated Q-Learning with \textit{Averaging of Q-Values} as the aggregation method}
    \label{fedavgq_qsum}
\end{figure}

\begin{figure}[!t]
    \centering
    \includegraphics[height=4.6cm, width=6cm]{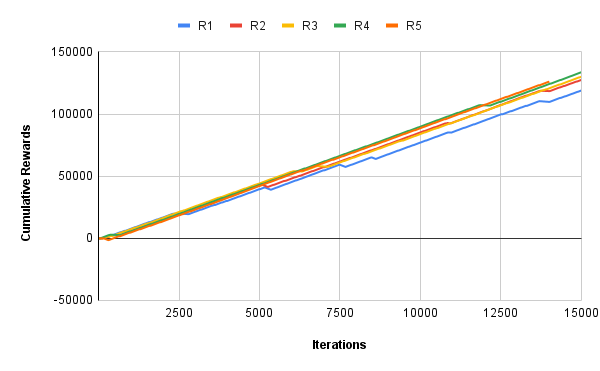}
    \caption{Cumulative Rewards over iterations of the Federated Q-Learning with \textit{Averaging of Q-Values} as the aggregation method}
    \label{fedavgq_rewards}
\end{figure}

In one experiment all the five robots, $ R_1$, $R_2$, $R_3$, $R_4$ and $R_5$, used Q-learning to hone their respective obstacle avoidance behaviours in their associated arenas while in the other, the robots used SARSA for the same. A mobile agent was programmed to carry the following functionalities in the order shown below:
\begin{enumerate}[label = (\alph*)]
  
\item  Access and carry the Q-table $Q^{(R_1)}$, generated by the first robot, $R_1$, after completion of 500 iterations. The iteration count was incremented every 64ms after which the Q-table was referred to find the next action to be taken.
\item  Carry $Q^{(R_1)}$ and migrate to the next connected \textit{Tartarus} platform. Access the Q-table, $Q^{(R_2)}$ of the associated robot ${R_2}$, and aggregate the contents of the two Q-tables to form a new Q-table, $Q^{(R_{1,2})}$.
\item  Carry $Q^{(R_{1,2})}$ and migrate to the next platform and repeat what was done in step (b) till it eventually reaches the last platform and generates the fully aggregated Q-table, $Q^{(R_{1,2,3,4,5})}$.
\item  The Q-table of ${R_{5}}$ was now updated to $Q^{(R_{1,2,3,4,5})}$. The mobile agent was then made to backtrack its path and while doing so update the locally available Q-tables, $Q^{(R_4)}$, $Q^{(R_3)}$, $Q^{(R_2)}$ and $Q^{(R_1)}$, respectively with the aggregated $Q^{(R_{1,2,3,4,5})}$ which it carried as its payload.
In this way, during the forward run, from the first robot ${R_{1}}$ to the last, ${R_{5}}$, the mobile agent performed the task of aggregating all the Q-tables, while in the backward run to the first robot, it shared the final aggregated Q-table. Since the mobile agent was programmed to continuously perform such migrations, after every 500 iterations of ${R_{1}}$, all robots were given sufficient chances to update/modify their respective  Q-tables based on their new experiences. 
\end{enumerate}

\begin{figure}[!t]
    \centering
    \includegraphics[height=4.6cm, width=6cm]{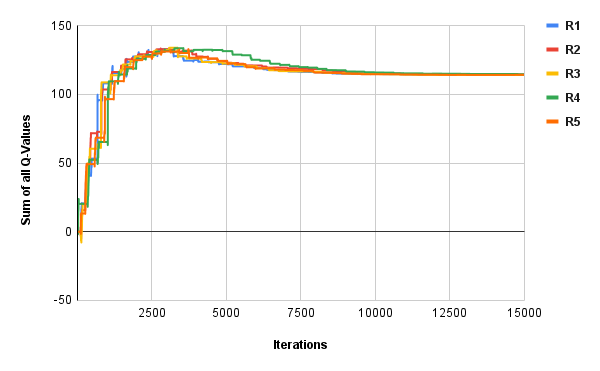}
    \caption{Sum of all Q-Values over iterations of the Federated Q-Learning with \textit{Maximum of Q-Values} as the aggregation method}
    \label{fedmaxq_qsum}
\end{figure}

\begin{figure}[!t]
    \centering
    \includegraphics[height=4.6cm, width=6cm]{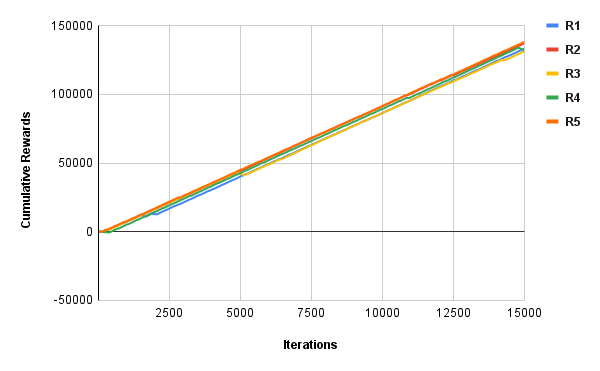}
    \caption{Cumulative Rewards over iterations of the Federated Q-Learning with  \textit{Maximum of Q-Values} as the aggregation method}
    \label{fedmaxq_rewards}
\end{figure}

\begin{figure}[!t]
    \centering
    \includegraphics[height=4.6cm, width=6cm]{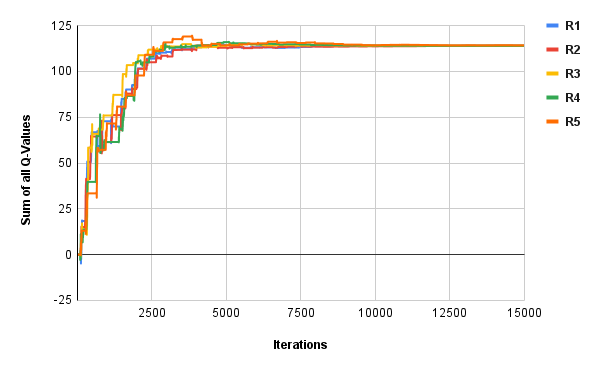}
    \caption{Sum of all Q-Values over iterations of the Federated SARSA with \textit{Averaging of Q-Values} as the aggregation method}
    \label{fedavgsarsa_qsum}
\end{figure}

\begin{figure}[!t]
    \centering
    \includegraphics[height=4.6cm, width=6cm]{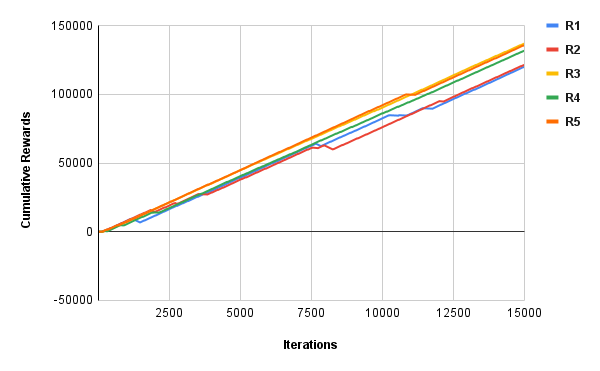}
    \caption{Cumulative Rewards over iterations of the Federated SARSA with \textit{Averaging of Q-Values} as the aggregation method}
    \label{fedavgsarsa_rewards}
\end{figure}

\begin{figure}[!t]
    \centering
    \includegraphics[height=4.6cm, width=6cm]{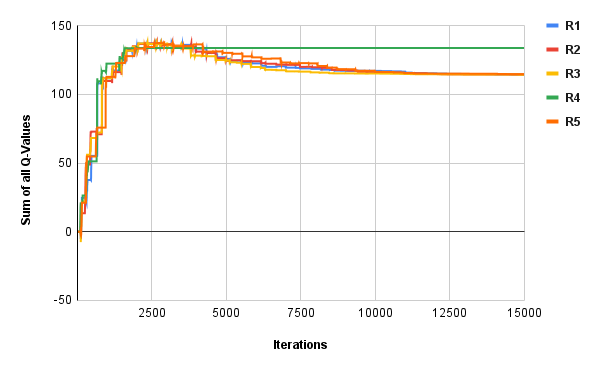}
    \caption{Sum of all Q-Values over iterations of the Federated SARSA with \textit{Maximum of Q-Values} as the aggregation method}
    \label{fedmaxsarsa_qsum}
\end{figure}

\begin{figure}[!t]
    \centering
    \includegraphics[height=4.6cm, width=6cm]{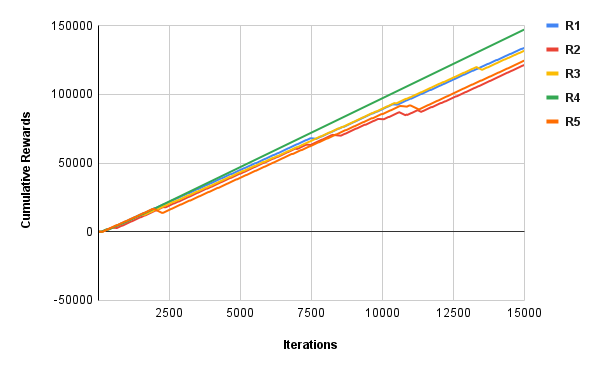}
    \caption{Cumulative Rewards over iterations of the Federated SARSA with \textit{Maximum of Q-Values} as the aggregation method}
    \label{fedmaxsarsa_rewards}
\end{figure}

The baseline graphs of the sum of all Q-values versus the number of iterations, for non-federated Q-learning and SARSA, wherein a single robot learns solo, have  been depicted in the graphs in the Fig. \ref{standaloneq} and \ref{standalonesarsa}. It can be seen that in the case of Q-learning, the robot, $R_1$, whose arena had just one solid block as the obstacle, learns faster than $R_4$. This is so because the latter had more obstacles to cope up with and thus, took more time. Eventually both the learning curve for both robots saturate. Similar trends can be observed in these robots when SARSA was used to learn the behaviour. 

The corresponding graphs of the sum of all Q-values, and the cumulative rewards obtained, versus the iterations, for both the aggregation methods used in conjunction with Q-learning and SARSA, for all the five robots,  ${R_{1}}$, ${R_{2}}$, ${R_{3}}$, ${R_{4}}$ and ${R_{5}}$, have been depicted in Figs. \ref{fedavgq_qsum} through \ref{fedmaxsarsa_rewards}. 
As can be observed in these graphs 
%
both the  sum of all Q-values and the corresponding cumulative rewards continue to increase with iterations indicating the constant increase in the learning of the behaviour by each robot. The aggregated Q-values eventually saturate, indicating the probable completion of learning across the different environments. Figs.\ref{fedmaxq_qsum} and \ref{fedmaxq_rewards} depict the graphs when the maximum of the Q-tables was used for aggregation in conjunction with Q-learning. Unlike the case of averaging (graph in Fig.\ref{fedavgq_qsum}), this graph exhibits a dip before saturating because the maximum of Q-values in each of the cells of the Q-tables populates the corresponding place in the aggregated Q-table.
Since environments of the robots are different, the move corresponding to the best/maximum Q-value of a robot need not be the best for another robot. When the corresponding action is taken by the other robot it is possible that it gets a lower reward and consequently, lower Q-value.
%
However, over time, as the learning improves, the Q-values of all robots, saturate in later stages. Here too, it may be observed that, the graph for $R_4$, differs from the others, due to the presence of more number of obstacles.

The graphs in Fig.\ref{fedavgsarsa_qsum}-\ref{fedmaxsarsa_rewards}, wherein the five robots use SARSA for learning, also seem to follow similar trends for both the aggregation methods. 
In all these cases, the baseline, i.e. when a single robot learns the behaviour solo, seems to saturate faster. This is obvious, as its learning mechanism needs to cater to only a single environment. It should be noted that if this robot were to be made to inhabit a new environment, it would take more time for it to learn, in contrast to those that have aggregated the learning through federation. FL, thus provides for a generic model catering to different environments.

\section{Discussions and Implications}
While this paper discusses the use of mobile agents to realize decentralized federated reinforcement learning, it is worthwhile to note that the same may be used in conjunction with other learning algorithms too. There is thus, no dependency on the learning technique used. Instead of RL, other learning techniques, such as Artificial Neural Networks (ANN) that use either deep or shallow architectures or neuro-evolution, could be used by the robots and the mobile agent could be made to aggregate the same to realize decentralized federated learning. Since the use of an agent provides for a mobile mechanism to access the learned contents, aggregate and share them across connected systems, in a decentralized manner, it unfolds a range of applications in multi-robot scenarios. 
The fact that both \textit{Tartarus} and Webots can be interfaced with real robots, this work can be used to realize decentralized federated learning with part simulated and part real robots in the loop. While the simulated ones can learn faster with no wear and tear, the real ones can generate learning models that cater to the more noisy real environments. Researchers can thus, use the best of both simulated and real worlds in their experiments. 
Though the work described herein, uses just one mobile agent and a single learning algorithm, it is now possible to implement decentralized FL using heterogeneous learning algorithms. In such a multi-robot scenario, robots within the same or different environments could use different learning algorithms to hone a behavior. The fact that \textit{Tartarus} supports multiple agents can be leveraged to realize this. Imagine a case where there are three robots inhabiting three different environments discussed earlier, with each robot using a different learning algorithm viz. $A_1$,$A_2$ and $A_3$, respectively. Three mobile agents capable of independently aggregating the models of the learning algorithms could be introduced into the system. An agent allocated to aggregate the models of A1 would contact only those robots that use this algorithm, aggregate them as per Algorithm 1, and send the final aggregated model to only these robots. This feature will allow researchers to use this work as a test bed and try dFL using different learning algorithms and then find the most suitable ones for a given environment. In addition, agents could be made to use different methods for aggregation. The best algorithm could eventually be made to replace other algorithms running on the the same or other robots. Since \textit{Tartarus} provides on-the-fly programming \cite{tartarus}, such new agents, with new aggregation methods, can be injected into the network without bringing down the overall networked system. It is also possible to add new robots using new algorithms on-the-fly into an environment.  
If the robot is placed in a new environment, the Q-values obtained from the federated learning setup will aid it to learn faster than in a case when the Q-values from a standalone environment are provided. This is because, the Q-values from the federated learning setup have adapted to various environments unlike the standalone one, resulting in the hastening of the learning process.

\section{Conclusions and Future Work}
This paper described a methodology to integrate mobile agent technology with simulated robots within Webots to eventually achieve decentralized federated reinforcement learning in multiple robots. Two variants of RL viz. Q-Learning and SARSA coupled with two aggregation methods were used to demonstrate the same. The paradigm running on machines running Webots and \textit{Tartarus} connected via LAN, was used to demonstrate its viability to support decentralization of FL. Robots were made to learn obstacle avoidance in different environments concurrently. Results obtained using two model aggregating methods, showed that decentralizing and implementation of federated learning in the domain of robotics is worthwhile since it can speed up or generalize the learning process across environments. 

It is important to note that though this paper exposes a mechanism to realize dFRL, the same may be used for other such learning techniques in the domain of robots. The use of a mobile agent has thus, yielded a mechanism to decentralize federated learning in the realm of connected multi-robot systems. With this mobile agent based decentralized FL framework, it will now be possible to make connected robots use different learning algorithms in a federated manner. In such scenarios, some robots could be using one type of learning method while others could be using another, resulting in a heterogeneous federated learning environment. Eventually, the robots could choose the better performing models and use the same for exhibiting the associated behaviours in their respective environments, using decentralized algorithm selection methods \cite{SEMWAL2020105920}. The use of multiple mobile agents could facilitate the concurrent learning of more than one behaviour in a decentralized and federated manner.
Further, a hybrid paradigm that combines the best of centralized and decentralized federated learning mechanisms may improve performance \cite{agrawal2020decentralizing} and reliability. Since Webots is software being used in education and research and also by the industry, the proposed work is bound to be a vital aid for users to build newer, enhanced and more flexible learning environments catering to multi-robot scenarios. Since agents in \textit{Tartarus} agents can be programmed on-the-fly \cite{tartarus}, the proposed work will help users to intervene during run-time and alter the flow and control of the learning mechanism. In addition, since both Webots and \textit{Tartarus} provide for control of real robots, the learned and aggregated models can now be passed on to physical robots for experimentation in the real world. Our future work includes decentralizing and federating other learning algorithms such as the deep Artificial Neural Networks, swarm algorithms, etc. in the domain of robotics and use the churned models in real robots.
\section*{Acknowledgments}
The first author would like to express his gratitude and appreciation for the facilities extended to him at the Robotics Lab. at the Dept. of Computer Science and Engg., Indian Institute of Technology Guwahati, India, and the Dept. of Computer Science and Engg., Federal Institute of Science and Technology, Angamaly, India for completion of the reported work.

\bibliographystyle{IEEEtran}
\bibliography{references}

\end{document}